\documentclass[11pt]{article}
\pdfoutput=1
\usepackage[final]{acl}

\usepackage{times}
\usepackage{latexsym}
\usepackage{booktabs}
\usepackage{algorithm}
\usepackage[noend]{algpseudocode}
\usepackage{enumitem}
\usepackage{tabularx}
\usepackage{ragged2e}
\usepackage{float}

\usepackage[T1]{fontenc}
\usepackage[utf8]{inputenc}

\usepackage{microtype}

\usepackage{inconsolata}

\usepackage{graphicx}

\usepackage{times}
\usepackage{soul}
\usepackage{url}
\usepackage[utf8]{inputenc}
\usepackage{graphicx}
\usepackage{amsmath}
\usepackage{amssymb}
\usepackage{amsthm}
\usepackage{booktabs}
\usepackage{setspace}
\usepackage{algorithm}
\usepackage{algpseudocode}
\usepackage{makecell}
\usepackage{tcolorbox}
\usepackage{courier} % optional, improves monospace font
\usepackage{listings}
\usepackage{stfloats}
\usepackage{xcolor}

\usepackage{array}
\usepackage{geometry}
\newtcolorbox{promptbox}{
  colback=gray!10,
  colframe=gray!50,
  boxrule=0.4pt,
  arc=2pt,
  left=4pt,
  right=4pt,
  top=4pt,
  bottom=4pt,
  fontupper=\ttfamily\small,
}
\title{Planning Agents on an Ego-Trip: Leveraging Hybrid Ego-Graph Ensembles for Improved Tool Retrieval in Enterprise Task Planning}

\author{
  \textbf{Sahil Bansal\thanks{The authors contributed equally to this work.}},
  \textbf{Sai Shruthi Sistla\footnotemark[1]},
  \textbf{Aarti Arikatala},
  \textbf{Sebastian Schreiber} \\
  SAP Labs, Palo Alto, CA, USA \\
  \texttt{\{%
  \href{mailto:sahil.bansal01@sap.com}{sahil.bansal01},\ 
  \href{mailto:sai.shruthi.sistla@sap.com}{sai.shruthi.sistla}%
} \\
\texttt{%
  \href{mailto:aarti.arikatala@sap.com}{aarti.arikatala},\ 
  \href{mailto:sebastian.schreiber@sap.com}{sebastian.schreiber}%
\}@sap.com}
}
\begin{document}
\maketitle
\begin{abstract}
Effective tool pre-selection via retrieval is essential for AI agents to select from a vast array of tools when identifying and planning actions in the context of complex user queries. Despite its central role in planning, this aspect remains underexplored in the literature. Traditional approaches rely primarily on similarities between user queries and tool descriptions, which significantly limits retrieval accuracy, specifically when handling multi-step user requests. To address these limitations, we propose a Knowledge Graph (KG)-based tool retrieval framework that captures the semantic relationships between tools and their functional dependencies. Our retrieval algorithm leverages ensembles of 1-hop ego tool graphs to model direct and indirect connections between tools, enabling more comprehensive and contextual tool selection for multi-step tasks. We evaluate our approach on a synthetically generated internal dataset across six defined user classes, extending previous work on coherent dialogue synthesis and tool retrieval benchmarks. Results demonstrate that our tool graph-based method achieves $91.85\%$ tool coverage on the micro-average \textit{CompleteRecall} metric, compared to $89.26\%$ for re-ranked semantic-lexical hybrid retrieval, the strongest non-KG baseline in our experiments. These findings support our hypothesis that the structural information modeled in the graph provides complementary signals to pure similarity matching, particularly for queries requiring sequential tool composition.
\end{abstract}

\section{Introduction}
Agentic systems powered by Large Language (LLMs) or Reasoning Models (LRMs) excel in planning and scheduling sub-tasks for complex requests \cite{kim2023llmcompiler,erdogan2025planandact,rawat2025preact}. While these systems effectively break down tasks into manageable logical sequences, evaluations have primarily focused on controlled settings with limited, well-defined tools that fit within a model's context window, such as web search, a calculator, etc.

Enterprise environments present greater challenges, with organizations relying on thousands of specialized tools with complex, often undocumented interdependencies. Conventional retrieval methods, particularly vector-based similarity search, frequently miss relevant tools, resulting in fragmented execution strategies. This limitation is especially critical in general-purpose agentic planning systems, where both initial and ongoing tool discovery form the foundation for effective task decomposition and execution.

Moreover, effective tool discovery is an essential prerequisite for meaningful task decomposition: Agents must first identify what capabilities are available before they can decide how to break down and solve a complex problem. Despite its centrality to real-world agentic long-horizon planning, this aspect remains underexplored in the existing literature \cite{huang2024understanding,wei2025plangenllms}.

We propose a structured semantic representation of enterprise tools using semi-structured data from tool descriptions and metadata. This approach produces a knowledge graph (KG) capturing relationships between tools, entities, and parameters, enabling better mapping of user queries to relevant tools. Our KG-enhanced retrieval mechanism uses neighborhood expansion to uncover implicit connections that traditional retrieval methods miss.

Our contributions are fourfold:
\begin{enumerate}
    \setlength\itemsep{0pt}
    \setlength\topsep{0pt}
    \item We propose a method to extract and model tool dependencies, facilitating tool trajectory discovery when explicit dependencies are missing.
    \item We introduce the \textit{Ensemble of Ego Graphs (EEG)} algorithm, which uses an ensemble of 1-hop ego tool graphs extracted from our overall tool graph via a hybrid node matching and neighborhood expansion technique to improve tool retrieval performance.
    \item Motivated by an internal analysis of enterprise user queries, we define six distinct query classes in the context of complex user queries. We present a novel pipeline for generating multi-step, multi-intent queries aligned with these classes. Our approach leverages tool dependency analysis, based on parameter–parameter relationships and LLM-inferred return parameter graphs, to identify feasible tool chains and to ensure the generated queries are coherent, contextually relevant, and faithful to their intended class.
    \item We evaluate the retrieval efficacy of our EEG algorithm on complex user queries generated using the above pipeline, using a \textit{CompleteRecall} metric~\cite{zhang-etal-2025-murre} specifically adapted to our tool retrieval setup, demonstrating significant improvements over baseline approaches.
\end{enumerate}

% These instructions are for authors submitting papers to *ACL conferences using \LaTeX. They are not self-contained. All authors must follow the general instructions for *ACL proceedings,\footnote{\url{http://acl-org.github.io/ACLPUB/formatting.html}} and this document contains additional instructions for the \LaTeX{} style files.

% The templates include the \LaTeX{} source of this document (\texttt{acl\_latex.tex}),
% the \LaTeX{} style file used to format it (\texttt{acl.sty}),
% an ACL bibliography style (\texttt{acl\_natbib.bst}),
% an example bibliography (\texttt{custom.bib}),
% and the bibliography for the ACL Anthology (\texttt{anthology.bib}).
\section{Related Work}
Recent works have explored the use of tool graphs for tool retrieval but exhibit notable limitations.

For instance, ControlLLM~\cite{liu2024controlllm} requires an adjacency matrix to construct the tool graph, assuming its structure is defined a priori. While offering a structured approach for tool selection and execution, this reliance on pre-specified connections limits its applicability in automation scenarios where tool relationships are unknown or evolving.

ToolNet~\cite{liu2024toolnet} employs graph-based iterative tool traversal similar to our approach. However, its graph construction depends on either extensive tool-use trajectories from code repositories and public datasets (unavailable for enterprise APIs) or LLM-generated trajectories, which frequently contain errors.

COLT~\cite{qu2024colt} employs a complex multi-step, multi-bipartite graph training process for transductive graph embeddings, but lacks automatic scene inference and clear application paths for novel queries, limiting its generalization capabilities.

Tool Graph Retriever~\cite{anonymous2024tool} extracts tool dependencies from documentation to create a graph. Unlike their approach, our method doesn't use a custom dependency identification model but instead leverages similarities between parameters and other entities, extracted via Open Information Extraction, to connect tools in a graph. 

Our approach also shares some similarities with ToolFlow~\cite{wang-etal-2025-toolflow}, which builds tool dependency graphs from documentation as well, but applies them to conversation generation rather than retrieval purposes. Building on the work presented in this paper, we further extended the methodology to synthetically generate complex, multi-step business queries in scenarios where output parameters are not available. This extension enables a more rigorous evaluation of the proposed graph-based retrieval mechanism.

Graph RAG-Tool Fusion~\cite{lumer2025graphragtoolfusion}, developed concurrently with our work, similarly combines vector retrieval with knowledge graph traversal but differs in two key ways: First, unlike their method that relies on synthetic tool graphs with well-defined dependencies suited to depth-first search, our approach semi-automatically converts real enterprise tools into graph representations, requiring only minimal manual input to tune prompts for accurate LLM interpretation of tool metadata and to define domain-specific ontologies and entity types that ensure semantic consistency during graph construction. Second, we identify ego-graph entry nodes using multiple vector representations rather than limiting ourselves to only semantic embeddings, making our method more practical for enterprise environments.

Another related approach incorporates tool knowledge directly into model parameters through training or fine-tuning, including multi-label classifiers~\cite{moon2024efficient} and LLMs~\cite{DBLP:conf/iclr/WangHJWB025}. However, these parameter-based methods are inadequate for dynamic enterprise environments with large, frequently changing tool ecosystems. For a comprehensive review of such approaches and their limitations, we refer readers to recent surveys like~\cite{Qu_2025}.

In summary, our approach combines automatic tool graph construction with multi-vector graph-retrieval mechanisms in a novel way, offering superior adaptability to dynamic enterprise environments compared to existing methods that either rely on synthetic graphs, lack contextual understanding, or cannot scale with frequently changing tool ecosystems.

\section{Methodology}
\subsection{System Overview}
Our proposed pipeline for tool retrieval is structured into two principal stages: an offline phase for building a structured Knowledge Graph (KG) from semi-structured tool documentation and metadata, and an online phase for retrieving relevant tools in response to user requests.

\begin{figure}[h!]
  \centering
  \includegraphics[width=\linewidth]{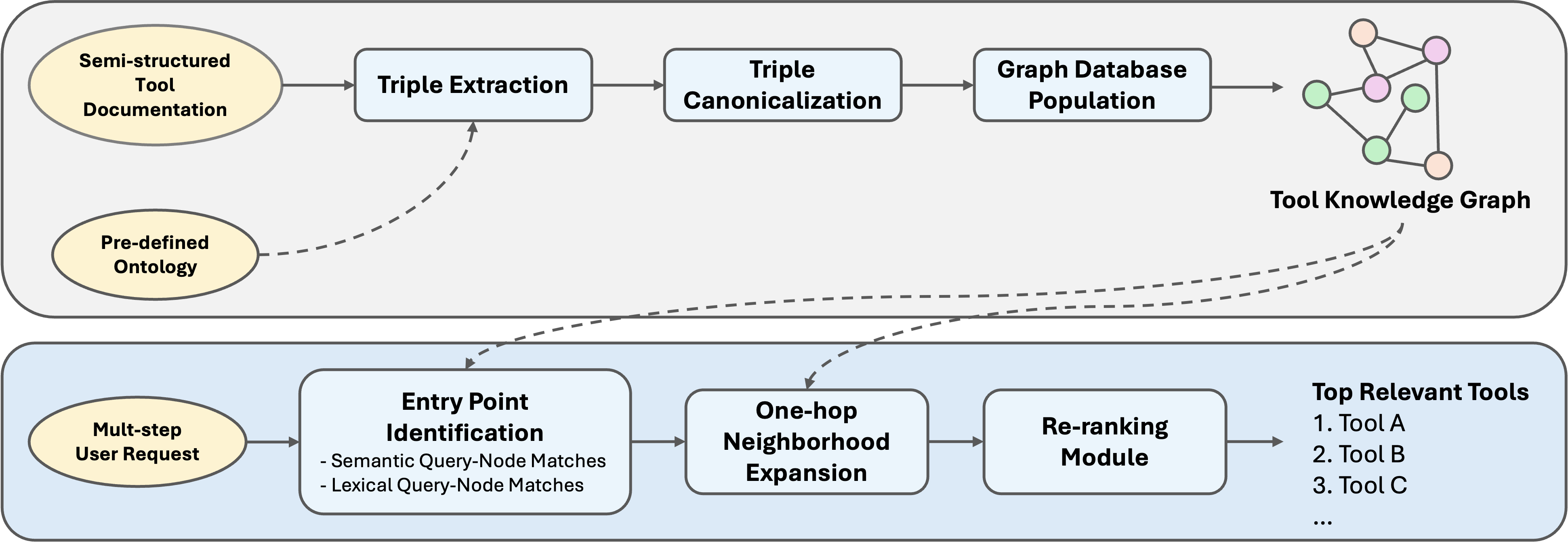}
  \caption{Tool Construction and Retrieval Pipeline}
  \label{fig:system_architecture}
\end{figure}

As illustrated in Figure~\ref{fig:system_architecture}, the offline phase begins with the ingestion of semi-structured tool documentation, which is processed by a triple extraction module to identify the relevant relational triples. Since naive extraction can result in an explosion of irrelevant or noisy triples, we guide the process using a pre-defined ontology that constrains and informs the extraction. The extracted triples are then passed through a triple canonicalization stage to ensure structural consistency and remove redundancies. The resulting canonical triples are used to populate a graph database, which instantiates the KG used for retrieval.

In the online phase, when a user query is received, our KG-based retrieval algorithm combines semantic query-tool node matching and textual query-entity node matching to identify entry points in the tool graph. One-hop neighborhood expansion then enriches the candidate tool set, followed by re-ranking to return the most relevant tools. The following subsections provide more details on these steps.

%  % forms the foundation of our knowledge graph.
\subsection{Knowledge Graph Construction} \label{sec:graph_construction}
\label{sec:kg_co}

Table~\ref{tab:tool-specs} introduces an example tool and its metadata, which we use as a running example throughout the methodology section.

\begin{table*}[t]
\centering
\small
% Adjust column padding for inner tables (optional)
\setlength{\tabcolsep}{4pt}
\begin{tabular}{@{} p{3cm} p{11cm} @{}}
\toprule
\textbf{Tool Title} & Send Task Deadline Reminder to Team Members \\
\midrule
\textbf{Description} & Sends an automated email reminder to team members responsible for tasks nearing their deadline. \\
\midrule
\textbf{Parameters} & 
\begin{tabular}[t]{@{} p{2.5cm} p{4.5cm} p{2.5cm} @{}}
\textbf{Name} & \textbf{Description} & \textbf{Type} \\
\midrule
priority\_filter & Filters for high-priority tasks requiring immediate attention. & \texttt{enum [High, Medium, Low]} \\
recipient\_list & Email addresses of team members assigned to tasks, extracted from \textit{task\_list\_upcoming\_deadlines}. & \texttt{list<user\_email>} \\
\end{tabular}
\\
\bottomrule
\end{tabular}
\caption{Sample tool with its metadata.}
\label{tab:tool-specs}
\end{table*}
% \begin{table*}[t]
% \centering
% \small
% \caption{Sample tool with its metadata.}
% \label{tab:tool-specs}

% \begin{tabular}{@{}p{3cm} p{12cm}@{}}
% \toprule
% \textbf{Tool Title} & Send Task Deadline Reminder to Team Members \\
% \midrule
% \textbf{Description} & Sends an automated email reminder to team members responsible for tasks nearing their deadline. \\
% \midrule
% \textbf{Parameters} &
% \begin{tabular}[t]{@{}p{3cm} p{6.5cm} p{2.5cm}@{}}
% \textbf{Name} & \textbf{Description} & \textbf{Type} \\
% \toprule
% priority\_filter & Filters for high-priority tasks requiring immediate attention. & \texttt{enum [High, Medium, Low]} \\
% recipient\_list & Email addresses of team members assigned to tasks, extracted from \textit{task\_list\_upcoming\_deadlines}. & \texttt{list<user\_email>} \\
% \end{tabular}
% \\
% \bottomrule
% \end{tabular}
% \end{table*}

To facilitate accurate and scalable retrieval of enterprise tools and their metadata, we construct a KG that captures both semantic and relational structures. This section details the four core components of our KG construction pipeline: ontology definition, triple extraction, triple canonicalization, and graph population. Our full graph construction algorithm in pseudo code is presented in Algorithm \ref{alg:graph_construction}.

\subsubsection{Ontology Definition}
To guide the extraction process and maintain a focused, manageable graph structure, we employ a predefined ontology tailored to our domain. This ontology defines:
\begin{itemize}
    \setlength\itemsep{0pt}
    \setlength\topsep{0pt}
    \item Entity Types: A restricted set of meaningful categories such as \textit{tool}, \textit{parameter}, \textit{line of business}, \textit{business object}, \textit{capability}, and \textit{department}. 
    \item Predicate Types: A curated list of relationships (e.g., \textit{has\_parameter, used\_by, assigned\_to, related\_to, contains}) that are relevant for retrieval and downstream reasoning tasks.
\end{itemize}

These types are directly aligned with the structure of enterprise task management tools. For instance, for the sample tool from Table~\ref{tab:tool-specs}, we derive:
\begin{itemize}
    \setlength\itemsep{0pt}
    \setlength\topsep{0pt}
    \item Entity:
    \begin{itemize}
    \setlength\itemsep{0pt}
    \setlength\topsep{0pt}
      \item \small \textit{type}: \texttt{tool}
      \item \small \textit{name}: \texttt{Send Task Deadline Reminder to Team Members}
    \end{itemize}
    \item Predicate-object pairs, such as:
    \begin{itemize}
    \setlength\itemsep{0pt}
    \setlength\topsep{0pt}
      \item \small \textit{has\_parameter}: \texttt{priority\_filter}
      \item \small \textit{has\_department}: \texttt{operations}
    \end{itemize}
\end{itemize}

By constraining both entity and relation types, we reduce graph noise, prevent combinatorial growth, and ensure that extracted triples directly support the system’s use-cases. The prompt which provides this guidance to the triple extraction LLM is shared in Figure \ref{fig:prompt}.

\subsubsection{Triple Extraction}
We extract structured semantic knowledge from semi-structured text sources by identifying relational triples of the form \textit{(subject, predicate, object)}. Each triple encodes a factual assertion about an entity and its relationship to another entity.
% \begin{itemize}
%     \item Subject: The source entity (e.g., \textit{a tool}).
%     \item Predicate: The relationship (e.g., \textit{has\_parameter}).
%     \item Object: The target entity or value (e.g., \textit{deadline\_range}).
% \end{itemize}
% \begin{itemize}[nosep, leftmargin=*, topsep=0pt]
%     \item Subject: The source entity (e.g., \textit{a tool}).
%     \item Predicate: The relationship (e.g., \textit{has\_parameter}).
%     \item Object: The target entity or value (e.g., \textit{deadline\_range}).
% \end{itemize}
\begin{itemize}
    \setlength\itemsep{0pt}
    \setlength\topsep{0pt}
    \item Subject: The source entity (e.g., a \textit{tool}).
    \item Predicate: The relationship (e.g., \textit{has\_parameter}).
    \item Object: The target entity or value (e.g., \textit{deadline\_range}).
\end{itemize}

To automate this process at scale, we leverage \texttt{GPT-4o}\footnote{\url{https://platform.openai.com/docs/models/gpt-4o}} for natural language understanding and relation extraction. The model is prompted with domain-specific instructions to identify and extract accurate and relevant triples from textual descriptions.
For example, for the tool in Table~\ref{tab:tool-specs}, we could extract the following triples:
\begin{itemize}
    \setlength\itemsep{0pt}
    \setlength\topsep{0pt}
    % \item (\texttt{Task}, \textit{categorized\_by}, \texttt{Priority Level})
    % \item (\texttt{Automated Reminder}, \textit{triggered\_by}, \texttt{Task Deadline})
    \item \small (\texttt{Task}, \textit{categorized\_by}, \texttt{Priority Level})
\item \small (\texttt{Automated Reminder}, \textit{triggered\_by}, \texttt{Task Deadline})
\end{itemize}

While some variations in entity or relation phrasing remain, LLM-based extraction mitigates many of the weaknesses of traditional approaches (e.g., rigid pattern-matching, limited semantic generalization) as shown in recent surveys of relation extraction using large language models~\cite{diaz2025survey,xu2024large}. Residual inconsistencies are addressed through a triple canonicalization step, which we describe next.

\subsubsection{Triple Canonicalization}
To enhance consistency and reduce redundancy in the KG, we perform normalization on both entities and predicates. This process involves two key steps:
\begin{itemize}
    \setlength\itemsep{0pt}
    \setlength\topsep{0pt}
    \item Entity Normalization: We unify different surface forms of the same entity into a single canonical representation. For instance, \textit{"Supplier", "supplier",} and \textit{"suppliers"} are all normalized to a single canonical node \textit{"supplier"}.
    \item Predicate Normalization: Semantically equivalent predicates are consolidated under a unified relation. For example, \textit{"works at"}, \textit{"employed by",} and \textit{"works for"} are normalized to a single canonical predicate \textit{"employed\_by"}.
\end{itemize}

This canonicalization step enhances graph quality by avoiding duplicate nodes and edges, which simplifies querying and downstream analysis.

\begin{algorithm}
\small
\caption{Knowledge Graph Construction\protect\footnotemark}
\label{alg:graph_construction}
\begin{algorithmic}[1]
\Require Tool documentation and metadata $D$, ontology $O$, domain-specific Open Information Extraction prompt $P$ 
\Ensure Knowledge Graph $KG$

\State $KG \gets$ initialize empty graph
\ForAll{tool\_doc in $D$}
    \State $triples \gets \Call{ExtractTriples}{tool\_doc, O}$
    \State $canonical\_triples \gets \Call{Canonicalize}{triples}$
    \ForAll{$(s, p, o)$ in $canonical\_triples$}
        \State \Call{AddNode}{$KG$, $s$, $metadata$}
        \State \Call{AddNode}{$KG$, $o$, $metadata$}
        \State \Call{AddEdge}{$KG$, $s$, $p$, $o$}
    \EndFor
\EndFor
\State \Return $KG$

\vspace{4pt}
\Function{ExtractTriples}{$doc, O$}
    \State \Return LLM($doc$, prompt=$P$, constrained\_by=$O$)
\EndFunction
\end{algorithmic}
\end{algorithm}
\footnotetext[2]{\textsuperscript{, 3}\;Monospaced text denotes function names in definitions and API calls; \textbf{bold} text marks language keywords; \emph{italics} mark variables, parameters, and arguments.}

\subsubsection{Graph Population}
Once the canonical triples are prepared, they are used to populate a graph database.
Each node in the graph is enriched with structured metadata, including \textit{"name", "id", "type"}, etc. 

\begin{algorithm}
\small
\caption{Ego Graph Tool Retrieval\protect\footnotemark}
\label{alg:tool_retrieval}
\begin{algorithmic}[1]
\Require User query $Q$, Knowledge Graph $KG$, embedding model $M$, reranker model $R$
\Ensure Ranked list of relevant tools $T$

\State $entrySem \gets \Call{MatchBySemanticSim}{Q, KG, M}$
\State $entryText \gets \Call{MatchByTextualSim}{Q, KG}$
\State $entryNodes \gets entrySem \cup entryText$
\State $candidateTools \gets \emptyset$

\ForAll{node in $entryNodes$}
    \State $egoGraph \gets \Call{OneHopNeighbors}{KG, node}$
    \State $tools \gets \Call{ExtractToolNodes}{egoGraph}$
    \State $candidateTools \gets candidateTools \cup tools$
\EndFor

\State $rankedTools \gets \Call{Rerank}{candidateTools, Q, R}$
\State \Return $\Call{TopK}{rankedTools, k = 10}$

\vspace{4pt}
\Function{MatchBySemanticSim}{$Q, KG, M$}
    \State $scoredNodes \gets$ \Call{EmbeddingSim}{$Q, KG, M$}
    \State \Return $\Call{TopK}{embeddingMatches, k = 10}$
\EndFunction

\vspace{4pt}
\Function{MatchByTextualSim}{$Q, KG$}
    \State \Return \Call{NodesWithExactNgramMatch}{$Q$, $KG$, $n\_max = 3$}
\EndFunction

\vspace{4pt}
\Function{Rerank}{$tools, Q, R$}
    \State $scoredTools \gets [\ ]$
    \ForAll{$tool$ in $tools$}
        \State $score \gets \Call{Reranker}{Q, tool, R}$
        \State Append $(tool, score)$ to $scoredTools$
    \EndFor
    \State \Return $tools$ sorted in descending order by $score$
\EndFunction
\end{algorithmic}
\end{algorithm}

\subsection{Ego Graph Retrieval}
\label{sec:ego_retrieval}
To retrieve the most relevant tools in response to a single or multi-step user query, we employ a custom ego-graph retrieval algorithm, described in Algorithm \ref{alg:tool_retrieval}, which consists of three main stages:
\subsubsection{Entry Point Identification in the Tool Graph}
\begin{itemize}
 \setlength\itemsep{0pt}
    \setlength\topsep{0pt}
    \item \textbf{Semantic Query-Node Matching:} 
    We embed the user query using OpenAI’s \texttt{text-3-embedding-large} embedding model and compute semantic similarity with all nodes in the graph. The \textit{top-10} most semantically similar nodes are selected as candidate entry points, ensuring alignment based on meaning.
    \item \textbf{Textual Entity-Node Matching:}
    We perform unigram, bigram, and trigram matching between the user query and the text associated with nodes in the graph. Any matching nodes are also considered as entry points, capturing more direct keyword-based connections.
\end{itemize}

\subsubsection{One-Hop Neighborhood Expansion for Tool Candidate Set Enrichment}
After identifying entry points, we execute a one-hop neighborhood expansion around each identified node, constructing an ensemble of ego tool graphs as previously described. This expansion process enriches our candidate set by incorporating all tool nodes directly connected to the entry points, thereby revealing contextually relevant tools that might otherwise remain undiscovered.

\subsubsection{Re-Ranking Retrieved Tools}
The \texttt{llama-3.2-nv-rerankqa-1b-v2}\footnote{\url{https://build.nvidia.com/nvidia/llama-3_2-nv-rerankqa-1b-v2/modelcard}} re-ranking model is used to re-rank the initially retrieved set of tools. The model takes as input the user query and each retrieved tool and outputs a relevance score for each. We retain the \textit{top-10} tools with the highest scores as the final output for a given user query.\\
An end-to-end example, along with a sample graph snippet, is provided in Appendix under section \ref{sec:complete_example}.

\section{Dataset Generation}

To evaluate our graph-based method, we require a tool retrieval benchmark suitable for enterprise use. Existing tool use benchmarks do not meet all these requirements, necessitating a custom dataset, cp. Table \ref{tab:tool_bench_comparison}.

\newcolumntype{Y}{>{\raggedright\arraybackslash\setlength{\parindent}{0pt}}X}

\subsection{User Query Classification in Enterprise Task-Oriented Systems}
\label{sec:user_query_classification_in_enterprise_task_oriented_systems}

Based on the requirements of our enterprise-oriented dialogue system, we have identified a taxonomy of query classes that reflect the diversity and complexity of real-world user requests. These classes help in understanding the structure, dependencies, and execution strategies required for accurate query interpretation and response generation. The classification is as follows:

\begin{itemize}
    \setlength\itemsep{0pt}
    \setlength\topsep{0pt}
    
    \item \textbf{Single-Intent Queries} involve only a single request with no additional steps, conditions, or dependencies, requiring direct execution.
    
    \item \textbf{Multi-Intent Queries} contain multiple independent requests that can be processed in any order or in parallel, with no logical dependencies between actions. 
    % \textit{Example:} ``Show my direct reports and display the weather forecast.''
    \item \textbf{Explicit Multi-Step Queries} include multiple actions where dependencies between steps are clearly stated in the query, requiring strict execution order.
    % \textit{Example:} ``Show me Jan's information and send him a spot award with a budget based on his career level.''
    \item \textbf{Implicit Multi-Step Queries} contain multiple actions where dependencies are implied rather than explicitly stated. The system must infer missing steps and their sequence before executing the main task. 
    % \textit{Example:} ``Assign Project Catalyst to the best-performing employee.'' 
    \item \textbf{Conditional Multi-Step Queries} explicitly state a condition that must be met before executing some of the actions involved. 
    % \textit{Example:} ``If quarterly revenue surpasses \$1M, schedule a bonus payout meeting.''
    % \item \textbf{Iterative Multi-Step Queries} involve repeated actions applied across multiple entities in a dataset, either targeting all records or a filtered subset. 
    % % \textit{Example:} ``Provide birthday details for all of Alex's peers.''
    \item \textbf{Information Retrieval + Multi-Intent Queries} combine general knowledge inquiry with personalized action, including both broad information requests and targeted instructions 
    % \textit{Example:} ``Explain the company’s bonus policies and display my latest performance review.''
     
    % \textit{Example:} ``I want to view the direct reports of John.''
\end{itemize}
An example for each type can be found in Table \ref{tab:complete_query_gen_results}. 

\subsection{Synthetic Multi-Step Query Generation}
\label{sec:query_generation_pipeline}

\begin{table*}[ht]
\centering
\resizebox{\textwidth}{!}{%
\begin{tabular}{lcccccccc}
\toprule
& \textbf{Ours} & \makecell{\textbf{ToolLinkOS}\\ \cite{lumer2025graphragtoolfusion}} & \makecell{\textbf{ToolSandbox} \\ \cite{lu-etal-2025-toolsandbox}} & \makecell{\textbf{ToolBench} \\ \cite{qin2023toolllm}} & \makecell{\textbf{ToolBank} \\ \cite{moon2024efficient}} & \makecell{\textbf{ToolRet} \\ \cite{shi-etal-2025-retrieval}}\\
\midrule
\textbf{Number of Tools} & 177 & 573 & 34 & 16,464 & 3,168 & 43,000 \\
\textbf{Number of Queries} & 503 & 1,569 & 1,032 & 126,486 & 163,000 & 7,600  \\
\textbf{Tool Dependencies} & \checkmark & \checkmark & \checkmark & x & x & x \\
\textbf{KG Schema} & \checkmark & \checkmark & x & x & x & x \\
\textbf{Complex Query Types} & \checkmark & x & x & x & x & x \\
\textbf{Business Tools} & \checkmark & x & x & x & x & x \\
\bottomrule
\end{tabular}
}
\caption{Comparison of our proposed business query dataset with other tool retrieval benchmarks.}
\label{tab:tool_bench_comparison}
\end{table*}

We introduce a structured \textit{Query Generation Pipeline} covering all the aforementioned query types. This pipeline comprises three key components—\textbf{Path Identification}, \textbf{Query Generation}, and \textbf{Query Validation}—that collectively synthesize realistic and semantically grounded user queries spanning the various user query classes.

\subsubsection{Path Identification}
\label{sec:path_identification}
This stage constructs meaningful tool chains by modeling both semantic and functional relationships using graph-based techniques. Specifically, we construct two graph structures to support diverse multi-step execution paths:

\begin{itemize}
    \setlength\itemsep{0pt}
    \setlength\topsep{0pt}
    \item \textbf{P-P Graph Construction:} Inspired by ToolFlow \cite{wang-etal-2025-toolflow}, this graph captures semantic proximity between tools based on cosine similarity of input parameter embeddings. Each node corresponds to a tool, and edges are established when similarity exceeds a predefined threshold, suggesting potential sequential usage or shared functional behavior. This structure enables efficient exploration of tool compositions for multi-step planning.
    \item \textbf{Inferred R-P Graph Construction:} We employ the \texttt{o3-mini}\footnote{\url{https://openai.com/index/openai-o3-mini/}} reasoning model to infer plausible output parameters for individual tools. These outputs are matched to tools accepting them as inputs, forming directed output-to-input edges annotated with confidence scores. Each tool sequence is passed through a validation step where we use \texttt{o3-mini} to reason where each generated sequence is valid. This graph reveals latent dependencies across tools, allowing for the construction of semantically valid but previously undocumented multi-tool flows.
\end{itemize}
Together, these graph structures enable robust path exploration for generating logically coherent multi-step queries.

\subsubsection{Query Generation}
\label{sec:query_generation}
Once valid tool sequences are identified, the pipeline synthesizes realistic user queries aligned with execution paths and class-specific semantics:
\begin{itemize}
    \setlength\itemsep{0pt}
    \setlength\topsep{0pt}
    \item Generates grammatically well-formed queries for each tool path.
    \item Adapts structure and phrasing to match one of six predefined user query classes, ensuring linguistic clarity and categorical separation.
    \item Instantiates abstract parameters with realistic sample values for contextual relevance.
    \item Promotes query diversity by varying linguistic style and avoiding repetitive formulations.
\end{itemize}
This step transforms tool logic into realistic language patterns, enabling robust evaluation of retrieval systems under multi-step user query conditions.
\subsubsection{Query Validation}
\label{sec:query_validation}

To ensure fidelity and structural correctness, the generated queries undergo systematic validation:
\begin{itemize}
    \setlength\itemsep{0pt}
    \setlength\topsep{0pt}
    \item \textbf{Class Validation:} Verifies that each query is properly classified according to its structural and semantic attributes.
    \item \textbf{Logical Sequence Verification:} Verifies that the tools used in a multi-step query are contextually compatible and collectively resolve the intended task. It checks whether each tool’s input and output logically align, preserving semantic coherence across the entire sequence.
    \item \textbf{Error Detection:} Identifies anomalies such as repeated tool references, incomplete requests, or incompatible parameter logic; such queries are flagged for exclusion.
\end{itemize}
These checks ensure the integrity of the synthetic dataset, enabling reliable evaluation of tool retrieval frameworks.

Detailed example prompts for R-P graph construction, query creation and validation are shared in Figures \ref{fig:output-parameter-prompt}--\ref{fig:scenario-sequencing-prompt} in Appendix \ref{sec:prompts}.

\begin{table}[H]
\centering
\small{
\begin{tabularx}{\linewidth}{
  >{\raggedright\arraybackslash}m{3.5cm}  % fixed width, ragged right, vertical centering
  Y                                        % no indent, ragged right, auto width
}
\toprule
\textbf{Query Type} & \textbf{Example} \\
\midrule

Multi-Intent & Can you show me the hire date of my manager, John, and then tell me which department he belongs to? \\
\midrule

Explicit Multi-Step & Can you show me the details of my expense with report ID \texttt{R1234} and then update the transaction amount to 500 with the currency code \texttt{USD}? \\
\midrule

Implicit Multi-Step & I need to adjust the transaction amount of my expense with report ID \texttt{R1234} to 500. \\
\midrule

Conditional Multi-Step & If the transaction date of my expense with report ID \texttt{R1234} is before 2022-01-01, update the transaction amount to 500. \\
\midrule

Information Retrieval + Multi-Intent & What's the current stock status? Also, adjust the product allocation profiles based on the stock information. \\
\bottomrule
\end{tabularx}
}
\caption{Representative Synthetic Queries Generated for Each Query Class}
\label{tab:complete_query_gen_results}
\end{table}

\subsection{Analysis of Generated Query Types}
\begin{figure}[h!]
  \centering
  \includegraphics[width=\linewidth]{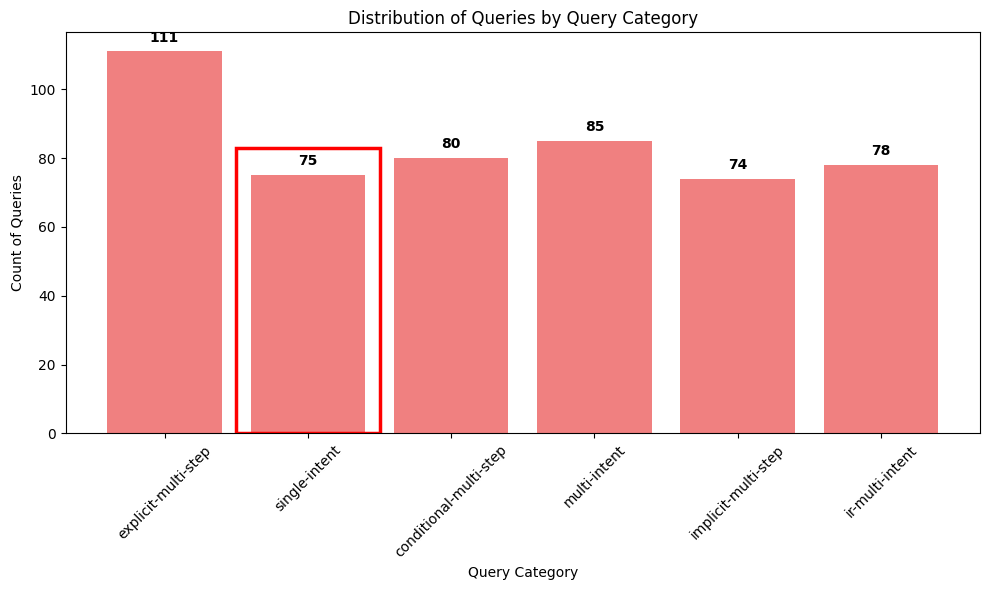}
  \caption{Query Distribution across identified query classes}
  \label{fig:query_distribution} 
\end{figure}

As a key outcome of our proposed pipeline, we generate a diverse set of synthetic queries spanning across five complex query classes. We analyze the distribution of these queries and provide qualitative examples to showcase how the pipeline captures the structural and semantic characteristics of real-world enterprise queries. As seen in Figure \ref{fig:query_distribution}, we categorized utterances into six distinct query classes based on intent category. The \textit{single-intent} category consists of real user queries sampled from production logs. The examples for the remaining query classes were synthetically generated using the method explained above to closely match this empirical distribution. Detailed examples for queries generated for each query type are present in Table \ref{tab:complete_query_gen_results}. This approach ensures our dataset reflects realistic usage patterns while enabling scalable coverage of more complex query types.

\section{Experimental Results}
\subsection{Experimental Setup}
We constructed our graph database using an internal dataset comprising semi-structured information extracted from several hundred enterprise tools within a large software platform. This dataset includes detailed descriptions, parameter specifications, and associated metadata, serving as the foundational data for the graph construction described in Section \ref{sec:graph_construction}.

To evaluate our system, we employed a separate dataset of synthetic user queries generated through the pipeline outlined in Section \ref{sec:query_generation_pipeline}. These queries were created by selecting targeted subsets of tools, formulating logical reasoning paths, and designing multi-step task-oriented queries. Each query was subsequently refined through a combination of automated validation and manual review to ensure high fidelity and practical alignment with real-world enterprise use cases.
% We evaluate our system using an internal dataset that encompasses enterprise tools across a large enterprise software platform. This dataset includes semi-structured information such as descriptions, parameter details, and metadata for several hundred tools. From this, we constructed our graph database as described in Section \ref{sec:graph_construction}.

% For evaluation, we use synthetic user queries generated using the query generation pipeline described in Section \ref{sec:query_generation_pipeline}. This process involves selecting specific tool subsets, constructing logical reasoning paths, and formulating multi-step queries. Each query undergoes automated and manual reviews to ensure realism and alignment with enterprise tasks.

\subsection{Evaluation Metrics}
We evaluate retrieval performance using the \textit{CompleteRecall} metric, defined formally as:
\[
\textit{CompleteRecall} = \frac{1}{|Q|} \sum_{q \in Q} \mathbf{1}(\textit{Recall@k}(q) = 1.0)
\]
where \( Q \) is the evaluation query set and \( \mathbf{1} \) is an indicator function that returns 1 if \textit{Recall@k} for a given query \( q \) is exactly $1.0$---meaning all required tools for that query are present within the \textit{top-k} retrieved items, with \textit{k} being the selected rank cutoff. This metric is tailored to planning systems where task breakdown depends on retrieving a complete set of expected tools. A comparable notion of complete recall has been used in prior work in the context of table retrieval~\cite{zhang-etal-2025-murre}.

 \subsection{Retrieval Results and Analysis}
To evaluate retrieval effectiveness, we compared four approaches: semantic retrieval, lexical retrieval (Okapi BM25\footnote{\url{https://pypi.org/project/rank-bm25/}}), hybrid retrieval, as well as our proposed KG-based retrieval method. Table~\ref{tab:complete_recall_results} summarizes the \textit{CompleteRecall} metric across all four approaches broken down by query type as well as aggregated averages. Comparable graph-based baselines are not available because no existing work provides enterprise-grade tools or real-world enterprise setups of this kind. As a result, direct empirical comparison is not feasible. 

Semantic retrieval using dense vector embeddings to capture the similarity between queries and tool descriptions achieved a micro-average \textit{CompleteRecall} of $59.84\%$ at $k=3$, $73.96\%$ at $k=5$, and $85.69\%$ at $k=10$, but struggled with nuanced queries requiring deeper contextual understanding.

% Semantic retrieval using dense vector embeddings to capture the similarity between queries and tool descriptions achieved $85.69\%$ \textit{CompleteRecall} but struggled with nuanced queries requiring deeper contextual understanding.

% For lexical retrieval, we evaluated Okapi BM25 with three tokenization strategies (simple whitespace split, regex-based tokenization with (\texttt{\textbackslash b\textbackslash w+\textbackslash b}), and SpaCy lemmatization\footnote{\url{https://spacy.io/api/lemmatizer}}) across two input types: tool description only and description plus title. Including titles generally improved performance, with regex-based tokenization reaching up to $75.33\%$ \textit{CompleteRecall} on the combined input. Despite this, lexical methods showed limitations in handling complex, multi-step queries common in enterprise workflows.
For lexical retrieval, we evaluated Okapi BM25 with three tokenization strategies: simple whitespace split, regex-based tokenization using \texttt{(\textbackslash b\textbackslash w+\textbackslash b)}, and SpaCy lemmatization\footnote{\url{https://spacy.io/api/lemmatizer}}. These were tested across two input types—tool description only, and description plus title. Including titles generally improved performance, with regex-based tokenization achieving micro-average \textit{CompleteRecall} of $48.31\%$ at $k=3$, $61.63\%$ at $k=5$, and $76.54\%$ at $k=10$ on the combined input. 
% Despite these gains, lexical methods showed limitations when handling complex, multi-step queries common in enterprise workflows.

% Semantic retrieval using dense vector embeddings to capture the similarity between queries and tool descriptions achieved $70.0\%$ \textit{CompleteRecall} but struggled with nuanced queries requiring deeper contextual understanding.

% For lexical retrieval, we evaluated Okapi BM25 with three tokenization strategies (simple whitespace split, regex-based tokenization with (\texttt{\textbackslash b\textbackslash w+\textbackslash b}), and SpaCy lemmatization\footnote{\url{https://spacy.io/api/lemmatizer}}) across two input types: tool description only and description plus title. Including titles generally improved performance, with regex-based tokenization reaching up to $70.0\%$ \textit{CompleteRecall} on the combined input. Despite this, lexical methods showed limitations in handling complex, multi-step queries common in enterprise workflows.
% Including titles generally improved performance, with SpaCy lemmatization achieving the highest score ($63.3\%$ \textit{CompleteRecall}) using only the descriptions.
Our hybrid baseline combines the \textit{top-10} results from both semantic and lexical retrieval approaches, and re-ranks them using \texttt{llama-3.2-nv-rerankqa-1b-v2}. This strategy achieves micro-average \textit{CompleteRecall} of $62.43\%$ at $k=3$, $78.93\%$ at $k=5$, and $89.26\%$ at $k=10$, outperforming the standalone methods.

% Our hybrid baseline combines the \textit{top-10} results from both the above approaches and re-ranks them using \texttt{llama-3.2-nv-rerankqa-1b-v2}, achieving $90.22\%$ \textit{CompleteRecall} — better than standalone methods but still lacking full contextual understanding.

\begin{figure}[h!]
  \centering
  \includegraphics[width=\linewidth]{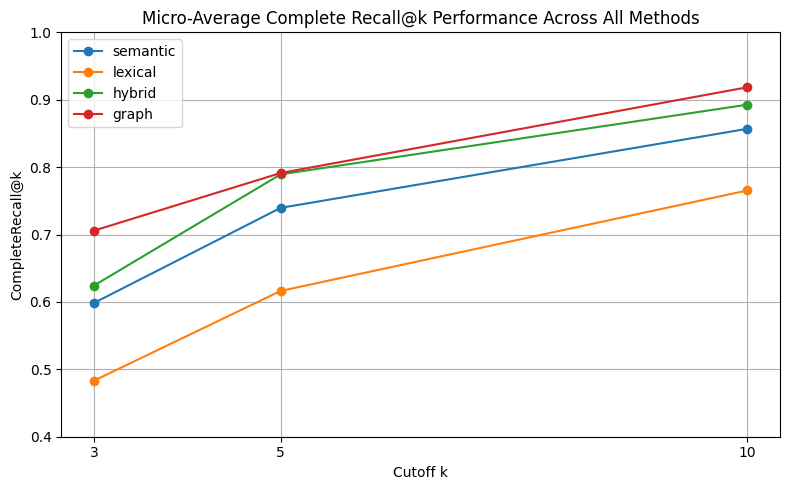}
  \caption{\textit{CompleteRecall@k} micro-averages for each retrieval method}
  \label{fig:tool_retrieval} 
\end{figure}
\begin{figure}[h!]
  \centering
  \includegraphics[width=0.48\textwidth]{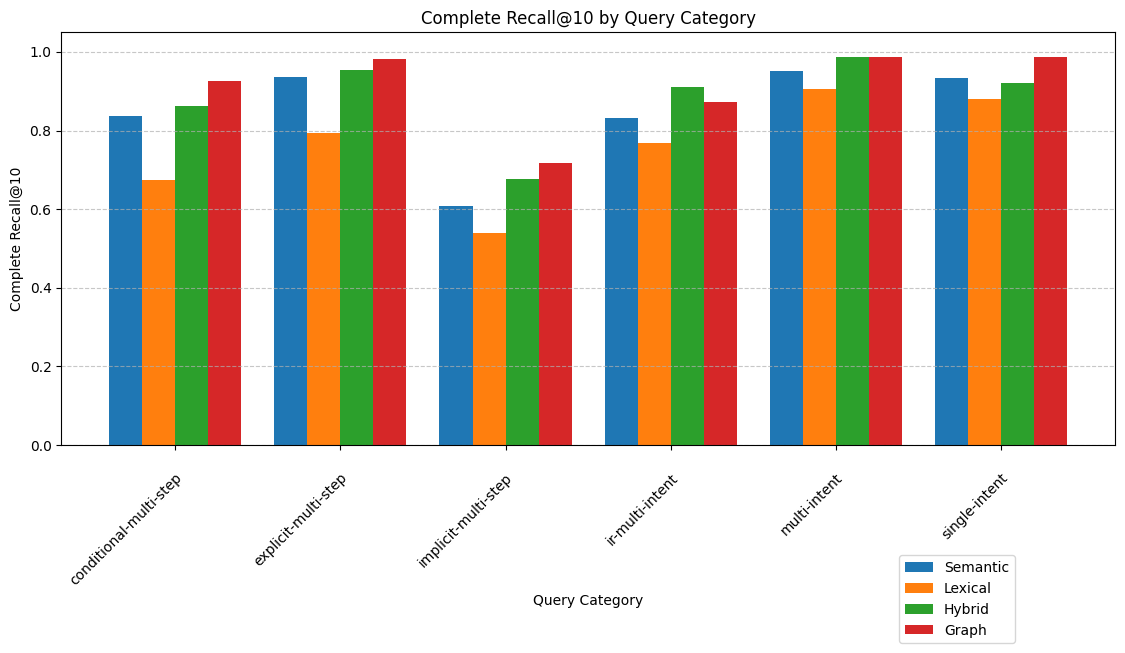}
  \caption{\textit{CompleteRecall@10} for each query type}
  \label{fig:tool_retrieval_category} 
\end{figure}

Our knowledge graph-based approach significantly outperformed all other methods, achieving micro-average \textit{CompleteRecall} scores of $70.58\%$ at $k=3$, $79.13\%$ at $k=5$, and $91.85\%$ at $k=10$, as shown in Figure~\ref{fig:tool_retrieval}. Figure~\ref{fig:tool_retrieval_category} highlights the most significant gains in complete recall within the conditional multi-step and implicit multi-step query categories. Appendix \ref{sec:add_results} provides further insights into our observations. We attribute these improvements to the KG’s ability to explicitly model semantic relationships among tools, enabling context-aware retrieval and revealing connections through shared functionalities and data dependencies.

% Our KG-based approach narrowly outperformed all other methods, achieving a  macro-average \textit{CompleteRecall} score of $90.88\%$ as seen in Figure \ref{fig:tool_retrieval}. As seen in Figure \ref{fig:tool_retrieval_category}, the largest improvements in complete recall are observed for the conditional multi-step and implicit multi-step query categories. We attribute this improvement to its ability to directly model semantic relationships between tools, enabling context-aware retrieval and uncovering connections through related functionalities and data dependencies. 

Some examples of its effectiveness:
\begin{itemize}
    \setlength\itemsep{0pt}
    \setlength\topsep{0pt}
    \item A query involving \textit{"budget planning"} and \textit{"project timelines"} retrieved interconnected tools such as \textit{financial forecasters} and \textit{scheduling applications} based on their shared association with time-based planning.
    \item Tools related to \textit{"user onboarding"} were retrieved as a group due to common entities such as \textit{authentication, documentation}, and \textit{workflow setup}, facilitating the discovery of a complete on-boarding toolkit.
\end{itemize}

These findings demonstrate that the proposed KG-based retrieval architecture enables more accurate and complete tool discovery, particularly in enterprise task planning environments that demand coordinated multi-tool workflows.
% \begin{table}
% \centering
% \caption{\textit{CompleteRecall} comparison across retrieval methods}
% \begin{tabular*}{\linewidth}{@{\extracolsep{\fill}} l c}
% \hline
% \textbf{Retrieval Method} & \textbf{CompleteRecall (\%)} \\
% \hline
% Semantic & 85.7 \\
% Lexical (Okapi BM25) & 74.6 \\
% Hybrid & 88.1 \\
% Knowledge Graph-based & \textbf{92.4} \\
% \hline
% \end{tabular*}
% \label{tab:complete_recall_results}
% \end{table}
% \begin{table}[h]
% \scriptsize
% \centering
% \begin{tabular*}{\linewidth}{@{\extracolsep{\fill}} lcccc}
% \hline
% \textbf{Query Category} & \textbf{Semantic} & \textbf{Lexical} & \textbf{Hybrid} & \textbf{Graph} \\
% % \textbf{} & \textbf{Recall@10 (\%)} & \textbf{Recall@10 (\%)} & \textbf{Recall@10 (\%)} & \textbf{Recall@10 (\%)} \\
% \hline
% conditional-multi-step & 83.75 & 65.00 & 82.50 & \textbf{92.50} \\
% explicit-multi-step    & 93.69 & 78.40 & 94.60 & \textbf{98.20} \\
% implicit-multi-step    & 60.81 & 54.10 & 68.90 & \textbf{71.62} \\
% ir-multi-intent        & 83.30 & 73.10 & 85.90 & \textbf{87.18} \\
% iterative-multi-step   & 91.00 & 71.00 & \textbf{93.00} & 86.00 \\
% multi-intent           & 95.20 & 90.60 & 98.80 & \textbf{98.82} \\
% single-intent          & 93.30 & 86.70 & 94.50 & \textbf{98.67} \\
% \hline
% \textbf{Aggregrated}            & 86.57 & 75.62 & 90.22 & \textbf{90.88} \\
% \hline
% \end{tabular*}
% \caption{Detailed Recall@10 (\%) by retrieval method across query categories.}
% \label{tab:complete_recall_results}
% \end{table}
\begin{table*}[t]
\scriptsize
\centering
\begin{tabular*}{\textwidth}{@{\extracolsep{\fill}} lcccccccccccc}
\hline
\textbf{Query Category} & \multicolumn{3}{c}{\textbf{Lexical}} & \multicolumn{3}{c}{\textbf{Semantic}} & \multicolumn{3}{c}{\textbf{Hybrid}} & \multicolumn{3}{c}{\textbf{Graph}} \\
& @3 & @5 & @10 & @3 & @5 & @10 & @3 & @5 & @10 & @3 & @5 & @10 \\
\hline
conditional-multi-step  &
42.50 & 52.50 & 67.50 & 
55.00 & 67.50 & 83.75 & 
\textbf{58.75} & \textbf{76.25} & 86.25 &
\textbf{58.75} & 72.50 & \textbf{92.50} \\
explicit-multi-step     &
43.24 & 58.56 & 79.28 &
71.17 & 84.68 & 93.69 &
71.17 & \textbf{90.09} & 95.50 &
\textbf{77.48} & 85.59 & \textbf{98.20} \\
implicit-multi-step     &
25.68 & 36.49 & 54.05 &
25.68 & 43.24 & 60.81 & 
32.43 & \textbf{51.35} & 67.57 &
\textbf{37.84} & 47.30 & \textbf{71.62} \\
ir-multi-intent         &
39.74 & 57.69 & 76.92 &
47.44 & 66.67 & 83.33 &
52.56 & \textbf{74.36} & \textbf{91.03} &
\textbf{62.82} & \textbf{74.36} & 87.18 \\
multi-intent            &
63.53 & 81.18 & 90.59 &
69.41 & 87.06 & 95.29 &
69.41 & 85.88 & \textbf{98.82} &
\textbf{90.59} & \textbf{94.12} & \textbf{98.82} \\
single-intent            &
76.00 & 82.67 & 88.00 &
84.00 & 88.00 & 92.00 &
85.33 & 89.33 & 92.00 &
\textbf{90.67} & \textbf{96.00} & \textbf{98.67} \\
\hline
\textbf{Micro-Average}            &
48.31 & 61.63 & 76.54 &
59.84 & 73.96 & 85.69 &
62.43 & 78.93 & 89.26 &
\textbf{70.58} & \textbf{79.13} & \textbf{91.85} \\
\hline
\end{tabular*}
\caption{\textit{CompleteRecall}@\{3, 5, 10\} (\%) across query categories and retrieval methods. Best @\{3, 5, 10\} score per row is highlighted in bold.}
\label{tab:complete_recall_results}
\end{table*}

% \newcolumntype{Y}{>{\raggedright\arraybackslash\setlength{\parindent}{0pt}}X}

% \begin{table}[htbp]
% \vspace*{-5em}
% \centering
% \caption{Representative Synthetic Queries Generated for Each Query Class}
% \label{tab:complete_query_gen_results}
% \begin{tabularx}{\linewidth}{
%   >{\raggedright\arraybackslash}m{3.5cm}  % fixed width, ragged right, vertical centering
%   Y                                        % no indent, ragged right, auto width
% }
% \toprule
% \textbf{Query Type} & \textbf{Example} \\
% \midrule

% Multi-Intent & Can you show me the hire date of my manager, John, and then tell me which department he belongs to? \\
% \midrule

% Explicit Multi-Step & Can you show me the details of my expense with report ID \texttt{R1234} and then update the transaction amount to 500 with the currency code \texttt{USD}? \\
% \midrule

% Implicit Multi-Step & I need to adjust the transaction amount of my expense with report ID \texttt{R1234} to 500. \\
% \midrule

% Conditional Multi-Step & If the transaction date of my expense with report ID \texttt{R1234} is before 2022-01-01, update the transaction amount to 500. \\
% \midrule

% Information Retrieval + Multi-Intent & What's the current stock status? Also, adjust the product allocation profiles based on the stock information. \\
% \bottomrule
% \end{tabularx}
% \end{table}

\section{Limitations}
While our approach improves tool retrieval performance for several complex query types, it has notable limitations. First, its effectiveness depends heavily on the quality and completeness of the underlying knowledge graph. Errors in triple extraction or missing tool relationships can considerably weaken results and create a lower bound on achievable performance, since inaccuracies in the extraction phase may propagate through later stages. Second, the framework may be less robust in domains with highly heterogeneous or sparsely described tools, reducing node-matching accuracy.

Surprisingly, for certain complex query types, traditional semantic and lexical-semantic hybrid retrieval methods outperformed our graph-based approach. This suggests that the additional structural complexity may not always provide benefits and warrants further investigation into when graph-based methods are most advantageous.

Finally, while we demonstrate strong results in our enterprise task domain, broader evaluation is needed to assess generalization across different domains and tool ecosystems.

\section{Conclusion and Future Work}
The task of efficiently exploring available tools, which is crucial for effective task decomposition during agentic planning, remains challenging for LLM-powered systems, particularly in enterprise environments with numerous tools that have complex and often undocumented interdependencies. Our research was guided by the hypothesis that explicitly modeling these relationships in a graph structure would enhance tool retrieval effectiveness.
To address the scarcity of complex queries in existing datasets, we propose a synthetic query generation pipeline that models tool dependencies through parameter-level connections, enabling the generation of realistic, multi-step queries.  

We present a systematic approach to transform enterprise tools into a coherent graph representation and introduc a novel \textit{Ensemble of Ego-Graphs (EEG)} retrieval framework that outperforms traditional baselines. Our results as shown in Table \ref{tab:complete_recall_results} support our hypothesis and establish a promising direction for improving tool retrieval in complex enterprise environments.  

Future research directions addressing some of the shortcomings we have identified include:
\begin{itemize}
    \setlength\itemsep{0pt}
    \setlength\topsep{0pt}
    \item Implementing a triple validation step to improve the quality of graph connections
    \item Adding additional dimensions to our dataset to go beyond the current defined classes and better capture the real-world variability and messiness of user queries.
    \item Making our dataset publicly available
    \item Exploring graph embedding techniques to complement our EEG retrieval algorithm 
    \item Developing methods to incorporate tool response information to enhance the tool graph's utility
    \item Evaluating and optimizing graph-retrieval latency to make it comparable with current retrieval mechanisms
    \item Introducing query augmentation strategies to inject realistic linguistic variability, such as ambiguity, underspecification, and incomplete phrasing to improve generalizability to real-world enterprise queries
    \item Incorporating inter-annotator agreement measures (e.g., Cohen’s Kappa) to evaluate consistency among experts and between experts and the automated validation pipeline to strengthen the reliability of evaluation outcomes
\end{itemize}
Through these efforts, we aim to further advance tool retrieval capabilities for enterprise applications.

\section{GenAI Usage Disclosure}
We employed ChatGPT and Claude to assist in rephrasing certain sections of the paper for improved clarity. All core content, including research design, data analysis, and result interpretation, was conducted without the aid of generative AI tools.
\bibliography{ACL_2025}

\clearpage

\appendix
\section{Complete Example}
\label{sec:complete_example}

\textbf{User Query:}
\noindent
\textit{Show me the details of all purchase order items with `pending' status.}

\vspace{6pt}
\noindent
\textbf{Target Tools:}
\begin{verbatim}
query purchase order item,
read purchase order item
\end{verbatim}

\noindent
\textbf{Entity-Node Matches:}
\begin{verbatim}
purchase order item, detail,
purchase order
\end{verbatim}

\noindent
\textbf{Top Relevant Tools Identified:}
\begin{verbatim}
read purchase order item,
query purchase order item,
show sale order query item,
read purchase requisition item,
query purchase order header,
read purchase order definition,
query purchase requisition item, 
show sale order read header,
read purchase requisition
definition
\end{verbatim}
\begin{figure}[h!]
  \centering
  \includegraphics[width=0.95\linewidth]{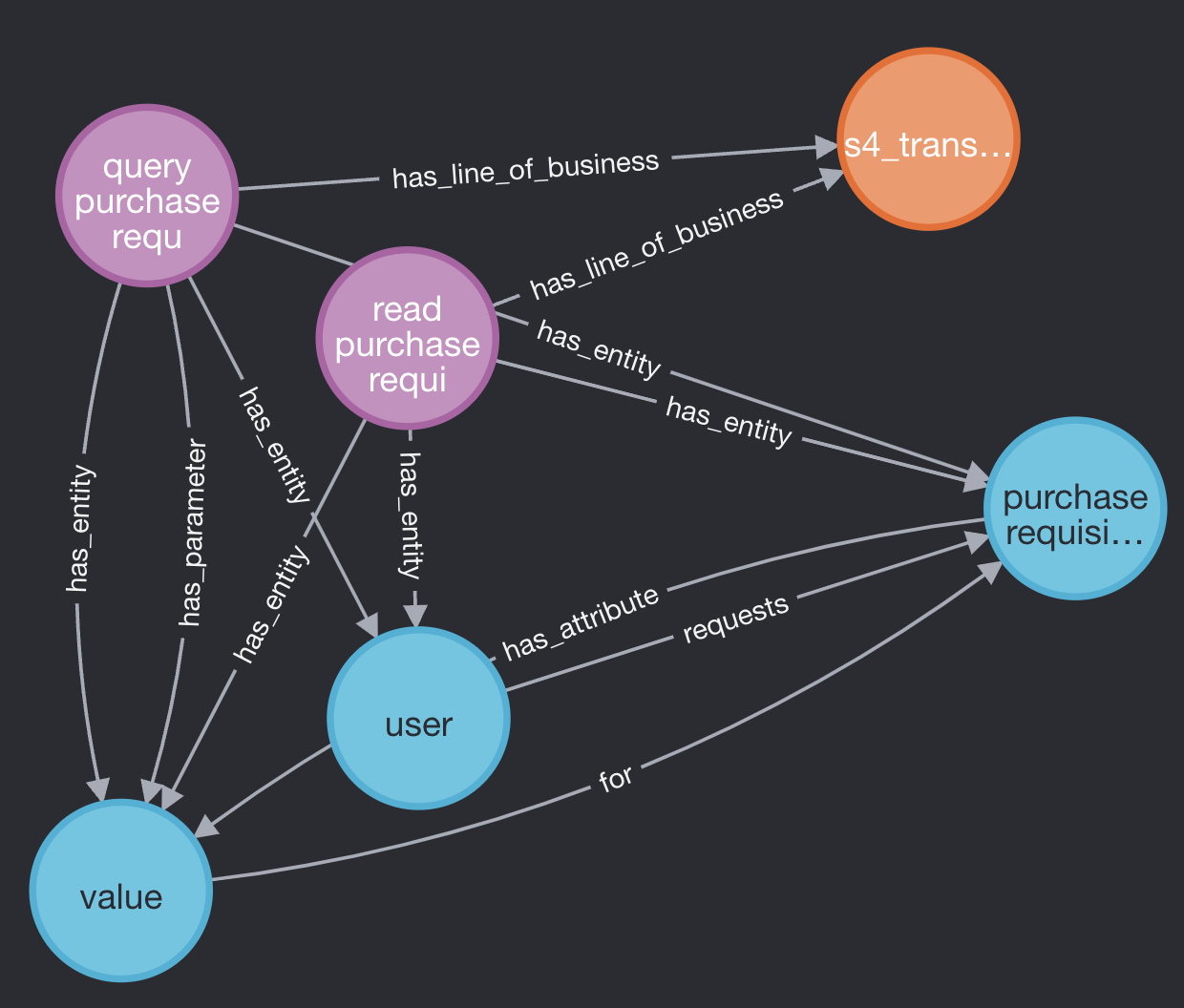}
  \caption{Subgraph to help illustrate the example in Appendix \ref{sec:complete_example}}
  \label{fig:example_graph_snippet}
\end{figure}
% \begin{figure}[h!]
%   \centering
%   \includegraphics[width=\linewidth]{example_graph_snippet.png}
%   \caption{Relevant subgraph to help illustrate the example}
%   \label{fig:example_graph_snippet}
% \end{figure}
\section{\textit{CompleteRecall@10} per Query Category}
\label{sec:add_results}

Figure \ref{fig:tool_retrieval_category} shows a histogram comparing the \textit{CompleteRecall@10} performance of each retrieval system across different query categories.

% \begin{figure}[h!]
%   \centering
%   \includegraphics[width=0.48\textwidth]{tool_retrieval_category.png}
%   \caption{\textit{CompleteRecall@10} for each query type}
%   \label{fig:tool_retrieval_category} 
% \end{figure}

\begin{itemize}
 \setlength\itemsep{0pt}
    \setlength\topsep{0pt}
\item For conditional multi-step, explicit multi-step, and implicit multi-step queries, the \textbf{graph} tool retrieval system performs best, indicating that the extracted triples effectively capture tool dependencies within the graph, enabling more accurate semantic reasoning and retrieval. 
% \item We also observe significant improvements for explicit multi-step queries. However, the hybrid model performs comparably well, as the dependencies in these queries are explicitly modeled and can be effectively decomposed semantically, even in the absence of underlying graph connections.
% \item For multi-intent, information retrieval multi-intent, and iterative multi-step queries, the graph model performs on par with the hybrid model.
\item For multi-intent, the graph model performs on par with the hybrid model.
\item For information retrieval multi-intent queries, the graph model performs under-par when compared with the hybrid model.
\item These results highlight that the graph model is especially effective for queries with implicit or complex dependencies that are not explicitly modeled, demonstrating the advantage of structured graph representations in capturing hidden relationships that improve retrieval performance.
\end{itemize}

\begin{figure*}[t]
\section{Prompts}
\label{sec:prompts}
\centering
\begin{minipage}{0.95\textwidth}
\begin{promptbox}
You are a top-tier natural language understanding expert, skilled in extracting triplets for Knowledge Graph construction.

\textbf{Objective:} Extract structured triplets (head–relationship–tail) from the given text to build a knowledge graph. The input is a scenario description from a digital assistant framework.

\textbf{Instructions:}

1. Read the provided text carefully.

2. Extract triplets in the format: \{ "head": ..., "tail": ..., "relationship": ... \}.
   
   - \textbf{head}: the main entity or concept performing an action or being described.
   
   - \textbf{relationship}: the action or relation connecting the head to the tail.
   
   - \textbf{tail}: the entity or concept receiving the action or being related.

3. Include default triplets based on specific patterns:
   
   - \texttt{has\_line\_of\_business}: between the scenario title and line of business.

   - \texttt{has\_entity}: between the scenario title and entities in the text.
   
   - \texttt{has\_parameter}: between the scenario title and listed parameters.

4. Also capture other relevant relationships, such as:
   \texttt{contains}, \texttt{related\_to}, \texttt{used\_for}, \texttt{has\_attribute}, \texttt{associated\_with},
   \texttt{managed\_by}, \texttt{part\_of}, \texttt{required\_for}, \texttt{depends\_on}, \texttt{produces},
   \texttt{receives\_from}, \texttt{involved\_in}, \texttt{reports\_to}, \texttt{responsible\_for}, \texttt{affects}, \texttt{includes}.

\textbf{Important Note:} For every "head" and "tail" entity in the triplets, include an additional \texttt{has\_entity} triplet linking it to the scenario title.

5. Output must be a valid JSON dictionary with this structure (and \textbf{no extra text}):
\begin{verbatim}
{
  "relationships": [
    {"head": "...", "tail": "...", "relationship": "..."},
    ...
  ]
}
\end{verbatim}

6. Example:

\textbf{Text:}
"The scenario titled 'Error Reporting' pertains to the Line of Business 'Finance'. The scenario description entails: 'As a user, I want to see a report of the top rejection errors for invoices.' The parameters are: error\_id, resolved\_number."

\textbf{Output:}
\begin{verbatim}
{
  "relationships": [
    {"head": "Error Reporting", "tail": "Finance",
     "relationship": "has_line_of_business"},
    {"head": "Error Reporting", "tail": "User", "relationship": "has_entity"},
    {"head": "Error Reporting", "tail": "Report",
     "relationship": "has_entity"},
    {"head": "Error Reporting", "tail": "Rejection error", 
     "relationship": "has_entity"},
    {"head": "Invoice", "tail": "Rejection error", "relationship": "contains"},
    {"head": "Report", "tail": "User", "relationship": "generated_for"},
    {"head": "Error Reporting", "tail": "error_id",
     "relationship": "has_parameter"},
    {"head": "Error Reporting", "tail": "resolved_number",
     "relationship": "has_parameter"}
  ]
}
\end{verbatim}

7. \textbf{Strict rules}:
- Do not duplicate triplets.
- Do not allow "head" and "tail" to be identical.
- The scenario title may only appear in \texttt{has\_line\_of\_business}, \texttt{has\_entity}, and \texttt{has\_parameter} triplets.
- Break complex sentences into simpler ideas to ensure accurate extraction.
- Maintain consistent naming of entities and relationships.

Follow the above instructions exactly, and output only a valid JSON dictionary.
\end{promptbox}
\end{minipage}
\caption{Prompt for domain-specific ontology-guided triple extraction from scenario descriptions.}
\label{fig:prompt}
\end{figure*}

\begin{figure*}[t]
\centering
\begin{minipage}{0.95\textwidth}
\begin{promptbox}
\textbf{System Prompt:}\\
You are an expert reasoning agent tasked with identifying output parameters in a business scenario.

\textbf{You will be provided with:} \texttt{(1)} a scenario name and description, \texttt{(2)} a list of input parameters and their descriptions, \texttt{(3)} a fixed list of \textbf{available parameters}, which are the only candidates you may choose from as outputs.

\textbf{Your Task:} Determine \textbf{which (if any)} of the available parameters are likely to be \textbf{outputs}—meaning they are \textbf{generated, updated, or returned} as a result of executing the scenario.

\texttt{-} Pay specific attention to the parameter descriptions when choosing a likely output.\\
\texttt{-} You may include up to 3 parameters.\\
\texttt{-} Only choose parameters that have \textbf{strong logical support} based on the scenario and inputs.\\
\texttt{-} If there is no clear evidence, return an empty list: \texttt{[]}.\\
\texttt{-} Do not guess or assume without justification—\textbf{precision is more important than recall}.

\textbf{Scoring Criteria:} For each selected parameter, provide a \textbf{confidence score} between 0 and 1, based on the following:

\texttt{-} \textbf{Relevance}: How directly the parameter aligns with the business goal or result described in the scenario.\\
\texttt{-} \textbf{Causality}: Whether the parameter is clearly generated or changed as a consequence of executing the scenario.\\
\texttt{-} \textbf{Clarity}: Whether the scenario description explicitly or implicitly implies this parameter is affected or produced.\\
\texttt{-} \textbf{Typical Usage}: Whether this parameter is commonly used as an output in similar scenarios or business processes.

\textbf{Scoring scale:}\\
\texttt{0.90–0.99}: Very strong evidence — directly and explicitly implied as an output; all four dimensions clearly supported.\\
\texttt{0.70–0.89}: Strong inference — not explicitly stated but logically follows from the scenario and typical practices.\\
\texttt{0.50–0.69}: Weak or partial evidence — some contextual hints or common patterns suggest it, but not clearly supported.\\
\texttt{< 0.50}: Do not include — insufficient support or speculative.

\textbf{Output Format:} Respond only with a JSON array, with \textbf{no markdown, no headings, and no surrounding text}. Each item must match the structure below and correspond exactly to entries in \texttt{available\_params}.

\textbf{Example:}
\begin{verbatim}
[
  {
    "parameter_name": "string (must match exactly from available_params)",
    "parameter_id": "string (must match exactly from available_params)",
    "confidence_score": float (0 to 1),
    "reasoning": "Short explanation (1–2 sentences max)."
  }
]
\end{verbatim}

\textbf{Notes:}\\
\texttt{-} Only select from the list: \texttt{\{available\_params\}}.\\
\texttt{-} Return \texttt{[]} if no likely output parameters.

\textbf{User Prompt:}
\begin{verbatim}
Scenario ID: {scenario_descriptions['scenario_id'][i]}
Scenario Name: {scenario_descriptions['joule_scenario_title_std'][i]}
Scenario Description: {scenario_descriptions['scenario_description'][i]}
Input Parameter Descriptions: {scenario_descriptions['Parameter_Info'][i]}
\end{verbatim}
\end{promptbox}
\end{minipage}
\caption{Prompt for identifying likely output parameters in business scenarios with confidence scoring.}
\label{fig:output-parameter-prompt}
\end{figure*}
\begin{figure*}[t]
\centering
\begin{minipage}{0.95\textwidth}
\begin{promptbox}
You are an intelligent query generation agent whose goal is to generate user queries using the logical function paths provided to you.

Here are the logical function paths:\{query\_generator\_input\}

Here are the different classes of user queries that have to be generated using these paths:

\textbf{1. Explicit Multi-Step User Queries}\\
\texttt{- Include multiple actions where each step explicitly depends on the completion of the previous one. }\\
\texttt{- Require a strict execution order, ensuring prior steps are processed before moving forward. Often use sequence-based phrasing such as "Show me X, then do Y."}\\
\texttt{Example: "Show me Jan's information and send him a spot award with a budget based on his career level."}

\textbf{2. Implicit Multi-Step User Queries}\\
\texttt{- Contain multiple actions, but the dependency between steps is implied rather than explicitly stated. The system must infer missing steps before executing the main task.}\\
\texttt{- User queries belonging to this class must not include sequencing phrases (e.g., "Do this, then do that"), conjunctions like "and" or "also" for distinct tasks, or conditional constructions (e.g., "If X happens, then do Y").}\\
\texttt{Example: "Send an email reminder to all suppliers invited to the Sapphire event." (Determining the list of invited suppliers is implicit.)}

\textbf{3. Conditional Multi-Step User Queries}\\
\texttt{- Depend on a condition being met before executing an action. Often use phrasing like "If X happens, do Y" or "Only if A is true, execute B." }\\
\texttt{- Require logical decision-making to ensure the correct steps are triggered.}\\
\texttt{Example: "Show me items related to GL 1234 if the account balance exceeds \$1M."}

\textbf{4. Multi-Intent User Queries}\\
\texttt{- Contain multiple independent requests that can be processed in any order or in parallel, with no logical dependencies between actions.}\\
\texttt{Example: "Show my direct reports and display the weather forecast."}

\textbf{5. Information Retrieval + Multi-Intent User Queries}\\
\texttt{- Combine a general knowledge inquiry with a personalized action. The informational query pertains to rules, definitions, or external facts, while the personal request focuses on user-specific data or tasks. Typically structured with both a broad question and a targeted action.}\\
\texttt{Example: "What is a spot award? Also, show me mine."}

\textbf{Instructions:}\\
\texttt{- Review the given logical path, including all functions, their purpose, descriptions, and input parameters.}\\
\texttt{- Generate one natural-sounding user query for each of the five classes based on the logical path.}\\
\texttt{- Ensure each query clearly reflects the intent of its respective class.}\\
\texttt{- Sound fluid, conversational, and human-like — avoid robotic or overly formal phrasing.}\\
\texttt{- Avoid internal domain-specific terminology and do not reuse exact words or phrases from function descriptions.}\\
\texttt{- Use realistic, fake values for at least one function’s input parameters (e.g. "location": "Chicago", "amount": 200) in the utterances.}\\
\texttt{- Make each query sound like something a real user might say in a relevant context.}\\
\texttt{- Verify that each user query distinctly and accurately reflects the intended class, ensuring no overlap or confusion between the different user intents.}

\textbf{Output Instructions:}\\
\texttt{- Provide the final output strictly in this format: \{format\_instructions\}}\\
\texttt{- Do not include extra text like "json" or "output" in the response.}
\end{promptbox}
\end{minipage}
\caption{Prompt for generating user queries across various multi-step classes using structured functional paths.}
\label{fig:user-query-generation-prompt}
\end{figure*}

\begin{figure*}[t]
\begin{minipage}{0.95\textwidth}
\vspace*{-28em}
\begin{promptbox}
\textbf{System Prompt:} \\
You are an expert in scenario analysis and workflow planning. Your task is to evaluate whether a sequence of two scenarios is valid based on their ability to follow one another in a logical, functional, and operational manner.

Each scenario includes a unique ID, a title, and a description of the actions or behaviors involved. The first scenario must be completed before the second one can logically occur, and the actions in the second scenario must be a valid continuation or follow-up to the first.

The scenarios must be part of a \textbf{multi-step process}, where the first scenario sets up a necessary context or action that the second can build upon. The scenarios cannot have distinct or unrelated intents. The second scenario must build upon the result or state created by the first. If the two scenarios are unrelated or do not form a cohesive multi-step action, the sequence should be considered invalid.

You will receive a list of two scenarios. Your task is to determine whether the second scenario can validly follow the first scenario in a multi-step process, based on the logical flow and dependencies between them.

\texttt{Return Format:} Provide a valid JSON dictionary with the following fields:
\begin{verbatim}
{
  "from_scenario_id": "string",
  "to_scenario_id": "string",
  "is_valid": true or false,
  "explanation": "Short rationale with example use case."
}
\end{verbatim}

\textbf{Notes:} \\
\texttt{- You must only select from this list of available parameters: \{available\_params\}} \\
\texttt{- Return an \textbf{empty list} (`[]`) if there are no likely output parameters.}

\vspace{1em}
\textbf{User Prompt:} \\
\texttt{Scenario ID: \{scenario\_descriptions['scenario\_id'][i]\}} \\
\texttt{Scenario Name: \{scenario\_descriptions['joule\_scenario\_title\_std'][i]\}} \\
\texttt{Scenario Description: \{scenario\_descriptions['scenario\_description'][i]\}} \\
\texttt{Input Parameter Descriptions: \{scenario\_descriptions['Parameter\_Info'][i]\}}
\end{promptbox}
\end{minipage}
\caption{Prompt for validating multi-step scenario transitions using structured logical analysis.}
\label{fig:scenario-sequencing-prompt}
\end{figure*}

\end{document}